\documentclass[sigconf]{acmart}

\usepackage{graphicx}
\usepackage{subcaption}
\usepackage{tabularx}
\usepackage{makecell}
\usepackage{multirow}
\usepackage{bbding}
\usepackage{tikz}
\usepackage{algorithm}
\usepackage{algorithmic}
\usepackage{booktabs}
\usepackage{threeparttable}
\def\halfcheckmark{\tikz\draw[scale=0.4,fill=black](0,.35) -- (.25,0) -- (1,.7) -- (.25,.15) -- cycle (0.75,0.2) -- (0.77,0.2)  -- (0.6,0.7) -- cycle;}

\begin{document}

%%
%% The "title" command has an optional parameter,
%% allowing the author to define a "short title" to be used in page headers.
\title{KnowMap: Efficient Knowledge-Driven Task Adaptation for LLMs}

\author{Kelin Fu}
\affiliation{%
  \institution{School of Computer Science}
  \city{Peking University}
  \country{China}
 }
\email{litble@stu.pku.edu.cn}

\author{Kaigui Bian}
\affiliation{%
  \institution{School of Computer Science}
  \city{Peking University}
  \country{China}
 }
\email{bkg@pku.edu.cn}

\begin{abstract}
While Large Language Models (LLMs) possess significant capabilities in open-world agent tasks, they also face challenges in rapidly adapting to new, specialized tasks due to their reliance on static pre-trained knowledge. Traditional methods such as fine-tuning are often costly, data-intensive, and may lead to "catastrophic forgetting." Therefore, we present KnowMap, a novel approach that dynamically constructs a knowledge base from environmental and experiential data. KnowMap fine-tunes a small knowledge-embedding model to equip a larger LLM with valuable task-specific knowledge. Our experiments on the ScienceWorld benchmark demonstrate 17.71\% improvement for the performance of \texttt{gpt-4-turbo} model. KnowMap not only provides an efficient and effective means for LLM task-adapting, but also highlights how integrating environmental and experiential knowledge can enhance LLMs' reasoning capabilities.
\end{abstract}

%%
%% This command processes the author and affiliation and title
%% information and builds the first part of the formatted document.
\maketitle

\section{INTRODUCTION}

% 备注：需要多介绍一下open-world text environment

\begin{figure}[t]
\centering
\includegraphics[width=0.99\linewidth]{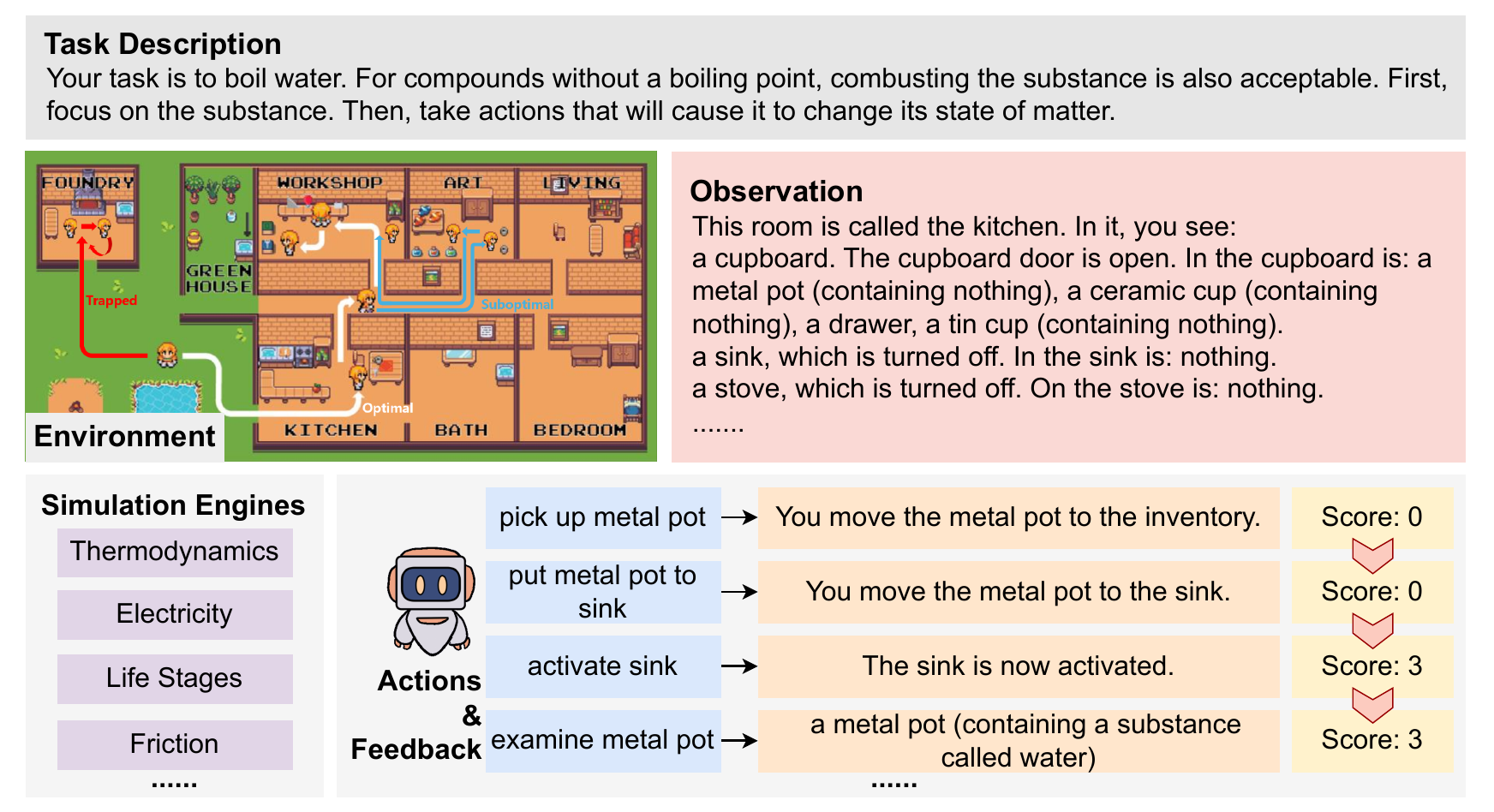}
\caption{Components of the embodied intelligence benchmark ScienceWorld. It uses text to convey observation information and action feedback to the agent, and the agent also uses text to submit its actions to change the environment and obtain feedback. This environment contains 10 interconnected locations, populated with up to 200 types of objects, and a action space with 25 high-level actions. The agent's each step have approximately 200k possible action-object combinations for selecting. And it supports multiple simulation engines that have an impact on the environmental state including thermodynamics, electricity, life stages and friction. It provides challenging tasks for agents and paves the way to real-life agents.}
\label{fig:sw_showcase}
\end{figure}

% 1. 提出问题：LLM 在agent领域的作用，以及面临的局限，如缺乏对特定领域的深度专业知识、难以快速适应新出现的开放世界中的各种新情况和新任务等。

Developing a generalist embodied AI (EAI) system capable of planning and acting within real-world environments remains a fundamental challenge in artificial intelligence. As an example, Fig.~\ref{fig:sw_showcase} shows an EAI benchmark in a text simulation environment, ScienceWorld~\cite{scienceworld}. The figure illustrates the types of tasks presented to the agent, the process of the agent submitting actions, and the resulting observations it receives from the environment. As an open-world environment with multiple simulation engines ranging from physical to chemical processes, ScienceWorld has over 200k possible action-object combinations for agents to select in each step, and countless environmental states. Such tasks challenges the agent's abilities in all aspects.

Nowadays, as large language models (LLMs) demonstrating remarkable capabilities in open-world tasks due to their proficiency in natural language understanding and complex reasoning, they have emerged as promising tools for tackling EAI problems. However, despite their versatility, LLMs typically lack deep domain-specific expertise and struggle to adapt rapidly to novel scenarios or emerging tasks in dynamic environments. This limitation stems from their reliance on static, pre-trained knowledge, which can become outdated or insufficient when confronted with highly specialized or evolving real-world challenges.

% 2. 现状不足：主流优化路径RL（计算成本高、奖励函数设计复杂），Seq2Seq SFT（需大规模标注数据、训练耗时长）和prompt（效果易饱和、依赖人工经验），微调大模型会削弱其通用性（灾难性遗忘），且资源消耗较大
Currently, three mainstream approaches are commonly employed to enhance LLM performance in specific tasks: (1) reinforcement learning (RL), (2) supervised fine-tuning (SFT), and (3) prompt engineering~\cite{prompt_engnieering}. However, RL is often hampered by high computational costs and the complexity of reward design. SFT, conversely, necessitates large-scale labeled data and lengthy training processes. Prompt engineering, while initially effective, frequently plateaus in performance and heavily depends on manual expertise. Furthermore, fine-tuning LLMs not only demands substantial computational resources but also carries the risk of catastrophic forgetting, where the model loses its general capabilities while adapting to new tasks. These inherent challenges render these methods unsuitable for all scenarios.

% 3. 解决方法1：强弱模型协同的新范式（可以补充更多知名一些的论文）
To mitigate these limitations, several studies, such as WESE~\cite{wese} and SwiftSage~\cite{swiftsage}, propose a collaborative paradigm that combines robust foundation models (general-purpose LLMs, strong models) with lightweight specialized models (fine-tuned for specific tasks, weak models). In this approach, the weak models handle simpler decisions, such as exploration tasks, while the strong models manage complex tasks like planning and state assessment. By decoupling task adaptation from general functionalities, these frameworks enable rapid adaptation to new scenarios.

% 4. 解决方法2：从文档RAG到结构性知识组织（补引用）
To further improve task adaptation, Retrieval-Augmented Generation (RAG) and knowledge-based methods have emerged as effective solutions. RAG integrates retrieval with generation, allowing models to quickly access relevant information from large corpora, thereby enhancing adaptation to new tasks without extensive retraining. Knowledge-based approaches advance this by organizing retrieved information into structured knowledge, facilitating more efficient reasoning and the application of general capabilities to new tasks. Interestingly, RAG also embodies the strong-weak model paradigm, where the retrieval component acts as a "weak" model to filter relevant information, and the generative model serves as a "strong" model to produce accurate outputs. Collectively, these methods offer a balanced approach to task adaptation, leveraging both efficiency and robustness to address the challenges of open-world environments.

% 5. KnowMap方法简介
Building upon these insights, we introduce KnowMap, a novel approach that dynamically constructs a knowledge base from environmental and experiential data. KnowMap then fine-tunes a knowledge-embedding model to supply a general decision model with task-specific knowledge. We evaluated KnowMap on ScienceWorld, a challenging open-world benchmark. Our experimental results demonstrate that fine-tuning an embedding model of approximately 0.56 billion parameters can improve the performance of \texttt{gpt-4-turbo} by 17.71\%. This approach not only presents an efficient strategy for large model task adaptation but also reveals the synergistic effect of knowledge derived from experience and the environment in enhancing knowledge encoding and large model reasoning.

\section{RELATED WORK}

% 还能多补一点知名的论文
\textbf{LLM agents for open-world tasks}. Open-world tasks present complex, interactive environments where agents must gather information through continuous interaction to achieve goals, prioritizing final rewards over process constraints. Early approaches, such as Chain-of-Thought (CoT) \cite{wei2022chain}, enhanced LLM reasoning by providing few-shot, step-by-step examples in prompts. Subsequent advancements include ReAct \cite{react}, which integrates explicit thought steps for action planning and reasoning, and Reflexion \cite{reflexion}, which enables iterative learning from past mistakes. Furthermore, works like LLM-planner \cite{song2023llm} and ReasonPlanner \cite{reasonplanner} address the construction of LLM agent decision-making scaffolds that balance long-term planning with short-term states, leading to more informed decisions. Inspired by these methodologies, KnowMap also constructs an agent decision-making scaffold with a planner and designs its experiential knowledge structure based on this framework's logic.

\textbf{Enhancing LLM with experiential knowledge}. To overcome LLMs' limitations in generalization, adaptability, and computational efficiency, prior research explores augmentation methods using experiential knowledge. LRLL \cite{LRLL} facilitates lifelong skill acquisition through dynamic memory and self-guided exploration, while ExpeL \cite{zhao2024expel} extracts reusable insights from interactions without fine-tuning, proving ideal for API-based LLMs. A case-based reasoning method \cite{atzeni2021case} boosts generalization in text games by repurposing past experiences, and Sub-goal Distillation \cite{subgoal_distillation} hierarchically transfers LLM knowledge to smaller models, reducing inference costs. Collectively, these approaches demonstrate how the structured reuse of experiential knowledge enables the development of adaptable, efficient, and generalizable agents. KnowMap extends this philosophy to open-world task adaptation.

\textbf{World models for LLM agents}. Recent research emphasizes leveraging structured world models, particularly knowledge graphs (KGs), to enhance LLM agents operating in complex, open-world environments. To address limitations such as greedy decisions and noisy information, WESE \cite{huang2024wese} decouples exploration and exploitation, using a cost-effective weak agent for global KG exploration to guide a stronger exploitation agent. Similarly, a study by AmmanaBrolu et al. \cite{ammanabrolu2021learning} explicitly frames the task as simultaneously predicting world state changes (represented as KG updates) and generating relevant actions via a transformer architecture solving a "Set of Sequences" problem, significantly outperforming prior methods. Focusing on reasoning over existing KGs, KG-Agent \cite{jiang2024kg} employs an autonomous LLM agent framework with a toolbox and KG executor, utilizing program language to structure multi-hop reasoning and effectively fine-tune smaller LLMs. Finally, AriGraph \cite{anokhin2024arigraph} integrates semantic and episodic memories into a structured graph (AriGraph) during exploration, demonstrating that this memory architecture, combined with planning, markedly improves complex task performance in interactive environments compared to unstructured memory methods and rivals dedicated KG techniques. Inspired by these contributions, KnowMap constructs its own environmental knowledge base and integrates it with an experiential knowledge base.

\section{METHOD}

\begin{figure*}[t]
\centering
\includegraphics[width=0.95\linewidth]{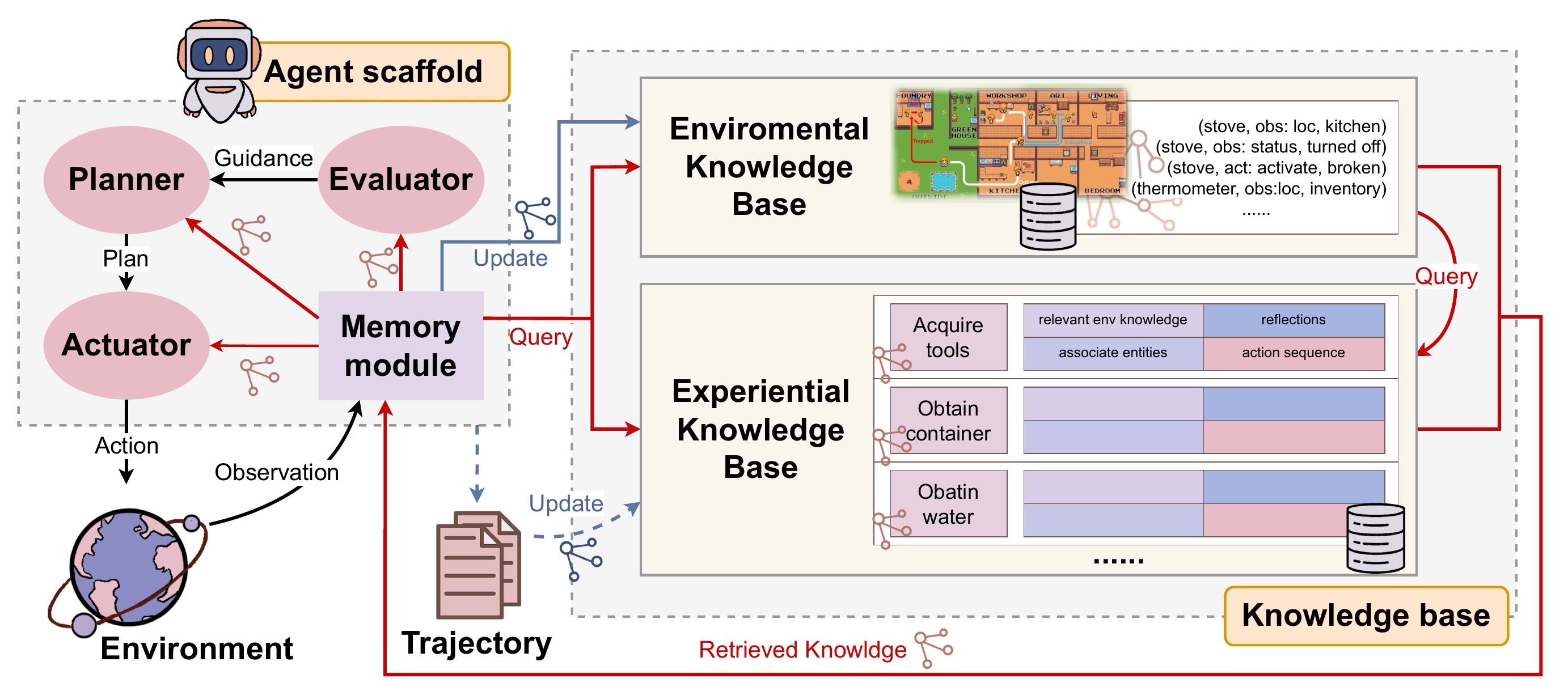}
\caption{KnowMap framework overview. Black arrows represent the agent's decision-making and interaction processes, red arrows indicate knowledge retrieval, and blue arrows denote knowledge updates.}
\label{fig:system}\end{figure*}

\subsection{Framework Overview}

The overall architecture of KnowMap is depicted as Fig.~\ref{fig:system}. It consists of two core components: the agent scaffold and the knowledge base. This section details the agent's descision-making mechanism and the knowledge base's retrieval/update processes.

\textbf{Agent scaffold}. The agent scaffold in KnowMap consists of three primary components: the planner, the actuator, and the evaluator. Specifically, the planner is responsible for formulating and adjusting the plans, while the actuator executes actions based on the formulated plans and interacts with the environment. The evaluator analyzes the execution status by examining the interaction history and current environmental observations, thereby providing feedback to guide the planner's adjustments.

In addition to these components, KnowMap's agent also includes a memory module designed to manage short-term memory for the current task. This memory module records all historical plans, actions, observations, and knowledge retrievals. It initiates queries and updates to the knowledge base and provides essential information to support decision-making by the planner, actuator, and evaluator.

\textbf{Query and retrieval of the knowledge base}. KnowMap’s knowledge base is divided into two primary types: environmental knowledge and experiential knowledge. A detailed introduction to the structure of the knowledge is provided in the Sec.~\ref{sec:knowledge_construction}. The knowledge retrieval process employs models based on BGE-M3 and the BGE-M3 Reranker~\cite{bgem3}. Initially, a combination of dense, sparse, and multi-vector retrieval methods is used to generate an initial set of candidate documents based on the query. The reranker is then employed to refine the selection, enhancing the accuracy and relevance of the final retrieved knowledge.

The retrieval of environmental knowledge primarily relies on task descriptions and plans formulated by the planner. The retrieval of experiential knowledge also utilizes this information, but additionally depends on the results obtained from the retrieval of environmental datasets.

\textbf{Update of the knowledge base}. The environmental knowledge base is updated throughout the task execution process. At the outset of a task, before the agent has gathered sufficient environmental observations, the knowledge base contains only information from the initial observations and some general knowledge. As the agent receives feedback from the environment in the form of observations or actions, the memory module extracts relevant knowledge, adds it to the knowledge base, and updates any outdated information.

In parallel, the experiential knowledge base is initially populated with knowledge derived from expert trajectories through LLMs. Additionally, the agent's own task trajectories, including those in which it completes sub-goals without achieving the overall task goal, are incorporated into the experiential knowledge base. This approach enables the agent to accumulate long-term experience and gradually progress toward achieving lifelong learning objectives.

\subsection{KnowMap's Knowledge Construction}
\label{sec:knowledge_construction}

\begin{figure}[htbp]
    \centering
    \begin{subfigure}[t]{0.49\textwidth}
        \centering
        \includegraphics[width=\linewidth]{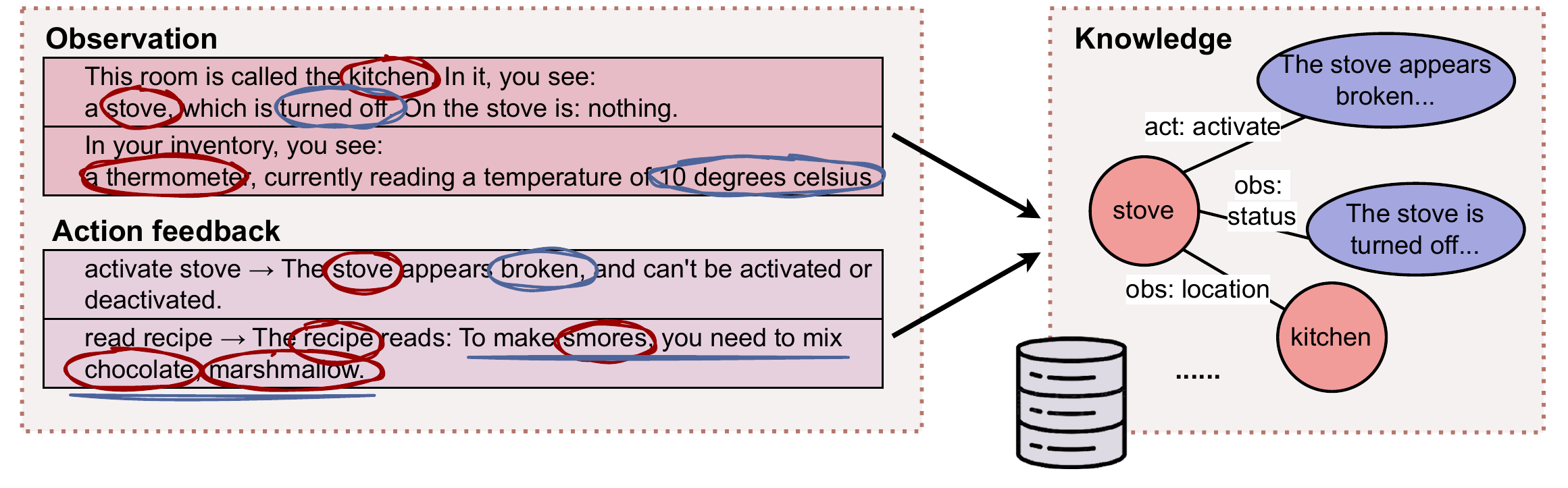}
        \caption{Example of constructing the environmental knowledge base in KnowMap. KnowMap acquires entities and their attributes in the environment through two pathways: observation and action feedback. It establishes different types of associations based on the methods of knowledge acquisition. When the associated objects of the same type of association for the same entity change, knowledge updates occur. In the figure, entities are marked with red circles, and attributes are marked with blue circles.}
        \label{fig:knowledge1}
    \end{subfigure}
    \hfill
    \begin{subfigure}[t]{0.49\textwidth}
        \centering
        \includegraphics[width=\linewidth]{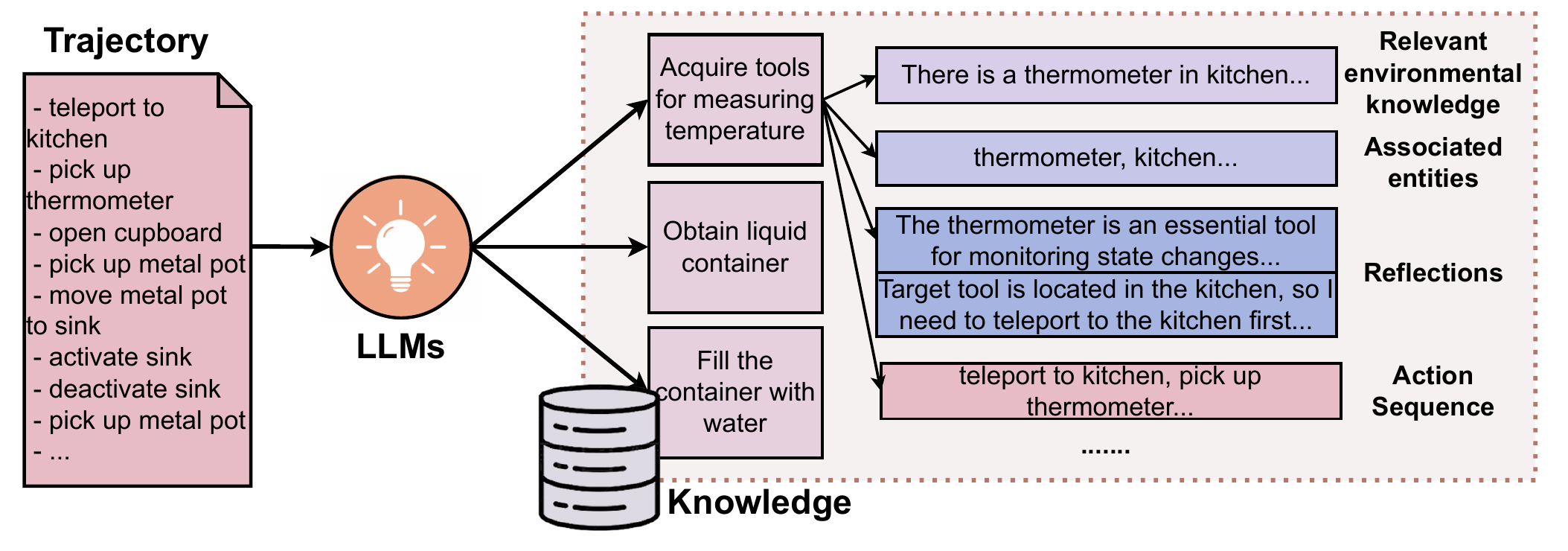}
        \caption{Example of constructing the experiential knowledge base in KnowMap. Through the LLMs, trajectories are decomposed into several sub-goals. Within each sub-goal, relevant environmental knowledge, associated entities, and reflections that support reasoning are extracted.}
        \label{fig:knowledge2}
    \end{subfigure}
    \caption{Construction of knowledge bases in KnowMap.}
\end{figure}

KnowMap's knowledge base incorporates two distinct knowledge bases, as illustrated in Fig. ~\ref{fig:knowledge1} and Fig. ~\ref{fig:knowledge2}: an environmental knowledge base and an experiential knowledge base. Each serves a specific purpose and exhibits a unique composition.

\textbf{Environmental knowledge}. The environmental knowledge base aims to maintain an accurate and current representation of all entities and their states within the agent's environment. The fundamental unit of knowledge within this base is a structured triple, formatted as \texttt{(entity, relation, entity/attribute)}. This triple captures a specific relationship (\texttt{relation}) linking an entity to either another entity or an attribute value.

To ensure knowledge currency, a replacement mechanism is employed: if a newly acquired triple shares identical entity and relation components with an existing triple, the new triple supersedes the original. For instance, an examine action on a thermometer yielding a new reading of 24°C would update the triple from \texttt{(thermometer, act: examine, 20°C)} to \texttt{(thermometer, act: examine, 24°C)}. Environmental knowledge acquisition occurs via observation or action feedback. 

Currently, the specific types of relations obtainable through each acquisition method are manually defined, though future work may investigate more flexible organizational approaches.

\textbf{Experiential knowledge}. The experiential knowledge base is designed to furnish the agent with resources for informed decision-making, including relevant examples, reusable experiences, and structured reasoning patterns to facilitate imitation and generalization. The core knowledge unit here is a sub-goal. Each sub-goal unit comprises the sub-goal name and associated summary information. 

This summary is extracted by decomposing expert trajectories into analyzable sub-goals using LLMs. For each sub-goal, the LLM extracts three key components: 

\begin{itemize}
    \item \textbf{Relevant environmental knowledge}: Identifies which environmental information influenced the expert’s decision-making in the given context.
    \item \textbf{Associated entities}: Highlights the key entities the expert focused on when making decisions.
    \item \textbf{Reflections}: Describes how the expert analyzed the environmental context to choose actions, and identifies generalizable experiences or decision patterns useful for future tasks.
\end{itemize}

% 记得添加引用
The prompt guiding this LLM-based trajectory analysis is detailed in the appendix. Notably, organizing trajectories generated by the KnowMap agent itself into this structured experiential knowledge is more straightforward than processing external expert demonstrations, due to full access to the agent's internal decision-making process.

\subsection{Fine-Tuning for KnowMap's Knowledge Embedding Model}

A trajectory $T$ is defined as a sequence of sub-goals $sg_1, \dots, sg_n$, where each sub-goal $sg_i$ comprises an action sequence $\text{traj}(sg_i) = {a_{i,1}, \dots, a_{i,m}}$ and corresponding experiential knowledge $\mathcal{K}_{\text{exp}}(sg_i)$. To enhance retrieval performance, we apply supervised fine-tuning (SFT) to the embedding model using trajectory-derived data. Each training instance consists of: \textit{(i)} a query $q$, \textit{(ii)} a positive knowledge sample $k_p$ and \textit{(iii)} a negative sample set $\mathcal{N} = {k_n^{(1)}, \dots, k_n^{(m)}}$.

We design distinct strategies for constructing training data from environmental and experiential knowledge bases.

\textbf{Fine-tuning data for the environmental Knowledge base}. The dataset $\mathcal{D}{\text{env}}$ is generated as:

\begin{enumerate}
    \item \textbf{Reproduce trajectories and collect knowledge}: We replicate actions from an expert trajectory $T_{\text{expert}}$ within the environment. Environmental knowledge $\mathcal{K}_{\text{env}}(sg_i)$ generated before initiating a specific sub-goal $sg_i$ is then added to the knowledge base.
    $$\mathcal{K}_{\text{env}}(sg_i) \leftarrow \text{EnvKnowledge}\left(\text{State}\left(\bigcup_{j=1}^{i-1} sg_j\right)\right)$$
    Here, $\text{State}\left(\bigcup_{j=1}^{i-1} sg_j\right)$ represents the environmental state resulting from the execution of all sub-goals prior to $sg_i$.
    \item \textbf{Identify interacted objects}: We scan the trajectory of the current sub-goal $sg_i$ to identify all objects $\mathcal{O}(sg_i)$ with which the expert has interacted.
    $$\mathcal{O}(sg_i) = \left\{o | o \text{ is interacted with in } sg_i \right\}$$
    \item \textbf{Formulate samples}: All knowledge associated with these identified objects is designated as positive samples ($\mathcal{P}_{\text{env}}(sg_i)$). Knowledge not related to these objects is designated as negative samples ($\mathcal{N}_{\text{env}}(sg_i)$).
    \begin{align*}
     \mathcal{P}_{\text{env}}(sg_i) &= \left\{k \in \mathcal{K}_{\text{env}}(sg_i) | \exists o \in \mathcal{O}(sg_i), o \text{ is related to } k\right\} \\
    \mathcal{N}_{\text{env}}(sg_i) &= \left\{k \in \mathcal{K}_{\text{env}}(sg_i) | \forall o \in \mathcal{O}(sg_i), o \text{ is not related to } k\right\}
    \end{align*}
    \item \textbf{Construct training data}: The training dataset for environmental knowledge, $\mathcal{D}_{\text{env}}$ is constructed by sampling an equal number of positive and negative samples for each sub-goal within the expert trajectories.
    $$\mathcal{D}_{\text{env}} = \left\{(sg_i, k_p, N_i) | k_p \in \mathcal{P}_{\text{env}}, N_i \subset \mathcal{N}_{\text{env}}(sg_i), \forall sg_i \in T_{expert}\right\}$$
\end{enumerate}

\textbf{Find-tuning data for the experiential knowledge base}. 
For the experiential knowledge base, we leverage similarities between sub-goal trajectories:

\begin{enumerate}
    \item \textbf{Calculate trajectory similarity}: We begin by calculating the cosine similarity between the embeddings of action trajectories of different sub-goal pairs using a general-purpose embedding model $\text{Emb}(\cdot)$.
    \[
    \text{sim}(sg_i, sg_j) = \frac{\text{Emb}(\text{traj}(sg_i)) \cdot \text{Emb}(\text{traj}(sg_j))}{\|\text{Emb}(\text{traj}(sg_i))\| \cdot \|\text{Emb}(\text{traj}(sg_j))\|}
    \]
    We then identify sub-goal pairs $(sg_i, sg_j)$ where their similarity exceeds a predefined \textbf{similarity threshold}, $\theta$:
    $$ \text{Pairs}_{\text{sim}} = \left\{ (sg_i, sg_j) \mid \text{Sim}(sg_i, sg_j) > \theta, i \neq j \right\} $$
    \item \textbf{Construct queries}: For each $sg_i$, a query $q_i$ is constructed by combining its associated environmental knowledge $\mathcal{K}_{\text{env}}(sg_i)$ with its trajectory $sg_i$:
    $$q_i = (K_E(sg_i), sg_i)$$
    \item \textbf{Identify positive samples}: The corresponding experiential knowledge $\mathcal{K}_{\text{exp}}(sg_i)$ from the matched sub-goal is used as a positive sample ($\mathcal{P}_{\text{exp}}(sg_i)$).
    $$
    \mathcal{P}_{\text{exp}}(sg_i) = \left\{k \in \mathcal{K}_{\text{exp}}(sg_i) | (sg_i, sg_j) \in \text{Pairs}_{\text{sim}} \right\}
    $$
    \item \textbf{Identify negative samples}: Randomly sampled experiential knowledge entries are used as negative samples.
    $$
    \mathcal{N}_{\text{exp}}(sg_i) = \left\{ k \in \mathcal{K}_{\text{exp}}(sg_k) \mid (sg_i, sg_k) \not \in \text{Pairs}_{\text{sim}}\right\}
    $$
    The training pairs for the experiential knowledge base would then be:
    $$\mathcal{D}_{\text{exp}} = \left\{ (q_i, k_p, N_i) \mid k_p \in \mathcal{P}_{\text{exp}}(sg_i),\ N_i \subset \mathcal{N}_{\text{exp}}(sg_i), \forall sg_i \in T_{expert} \right\}$$
\end{enumerate}

\textbf{SFT Optimization Objective}. Finally, to train the embedding model $\phi$  to discriminate the positive samples from the negative ones, we minimize the InfoNCE loss with temperature $\tau$:

\begin{multline*}
\mathcal{L} = -\frac{1}{|\mathcal{D}|} 
\sum_{\substack{(q,k_p,\mathcal{N}) \\ \in\, \mathcal{D}}} 
\log \left( 
\frac{e^{\phi(q)^\top \phi(k_p) / \tau}}{
e^{\phi(q)^\top \phi(k_p) / \tau} + 
\sum\limits_{k_n \in \mathcal{N}} e^{\phi(q)^\top \phi(k_n) / \tau}}
\right)
\end{multline*}

where $\mathcal{D} = \mathcal{D}{\text{env}} \cup \mathcal{D}{\text{exp}}$ and $|\mathcal{N}| = m$ negative samples per instance. This objective maximizes the similarity between queries and their positive samples while pushing away multiple negative samples simultaneously.

%\subsection{Synergy between hierarchical knowledge}.

\section{EXPERIMENTS}

\subsection{Evaluation Setup}

\textbf{Benchmark}. We select ScienceWorld~\cite{scienceworld}, an embodied AI benchmark based on a text simulation environment, as our experimental benchmark. ScienceWorld comprises 30 types of tasks derived from the grade school science curriculum, involving the exploration and reasoning of scientific principles such as state changes, conductivity, life cycles, genetics, and more. Each task includes over 100 variations to prevent overfitting. The rich diversity of task types, environmental state space, and action space makes ScienceWorld a challenging benchmark. The task score in ScienceWorld is the primary metric we use for comparison, representing the degree to which the agent completes the task. For instance, if an agent's final score is 60, it means that the agent has completed 60\% of the task before exhausting the action count or terminating the task due to incorrect actions.

\textbf{LLM backbones}. To verify the effectiveness of KnowMap in task adaptation, we avoid selecting backbone models that achieve extremely high zero-shot scores on the ScienceWorld benchmark. We primarily use \texttt{gpt-4-turbo} as the backbone in our experiments, along with \texttt{gpt-4o-mini} and \texttt{DeepSeek-V3-241226}.

\textbf{Baselines}. To evaluate KnowMap's performance, we compared it against several established baselines on the ScienceWorld benchmark. These baselines represent diverse approaches to task adaptation and agentic behavior in LLMs:

\begin{itemize}
    \item \textbf{SayCan}~\cite{saycan}: This method enables robots to execute complex natural language instructions by combining an LLM's world knowledge ("Say") with robotic affordances ("Can"), scoring the utility and feasibility of skills.
    \item \textbf{ReAct}~\cite{react}: ReAct is a framework that interleaves "Reasoning" (LLM-generated thoughts) and "Acting" (task-specific actions) to improve planning, handle exceptions, and interact with environments, overcoming limitations of reasoning-only approaches.
    \item \textbf{Reflexion}~\cite{reflexion}: Reflexion reinforces language agents through linguistic feedback. It allows LLM agents to reflect on feedback and store these reflections in episodic memory, enabling them to learn from mistakes and refine decision-making over time.
    \item \textbf{SwiftSage}~\cite{swiftsage}: This approach combines strong and weak models, using fine-tuned smaller models for action generation and larger models for critical action planning and evaluation.
    \item \textbf{Sub-goal Distillation}~\cite{subgoal_distillation}: This framework improves small-model performance by applying knowledge distillation to both the planner and actuator.
    \item \textbf{ReasonPlanner}~\cite{reasonplanner}: ReasonPlanner is an agent framework that uses a temporal knowledge graph for world modeling and planning, and a natural language actor-critic module for environment interaction.
\end{itemize}

All listed baselines, with the exception of Sub-goal Distillation, directly utilize large language models. For consistency, all reported results for these baselines were obtained using the \texttt{gpt-4-turbo} as the LLM backbone.

% 除了柱状图还能画点啥图吗
\subsection{Results}

% 备注2：需要再检查一下SwiftSage和Sub-goal Distillation使用的抽样方式（他们不会跑的是全集吧？）。
% 备注3：SwiftSage怎么这么高，要不删了算了（捂脸）
\begin{table}[t]
\centering
\begin{threeparttable}
\resizebox{0.49\textwidth}{!}{
  \begin{tabular}{c | c c | c}
    \toprule 
     Method & Using LLM & Tuning & Score (\%) \\ \midrule
    \rowcolor{gray!20}SayCan$^*$ & \Checkmark & \XSolidBrush & 33.82 \\
    ReAct$^*$ & \Checkmark & \XSolidBrush & 36.43 \\
    \rowcolor{gray!20}Reflexion$^*$ & \Checkmark & \XSolidBrush & 45.34 \\
    SwiftSage$^*$ & \Checkmark & \Checkmark & 84.68 \\
    \rowcolor{gray!20}Sub-goal Distillation$^\dagger$ & \XSolidBrush & \Checkmark & 65.43 \\
    ReasonPlanner$^\ddagger$ & \Checkmark & \XSolidBrush & 65.06 \\ \midrule
    \rowcolor{gray!20}KnowMap w/o embedder tuning & \Checkmark & \XSolidBrush & 66.90 \\
    KnowMap & \Checkmark & \halfcheckmark &  76.25\\
    \bottomrule 
   \end{tabular}
   }
   \begin{tablenotes} % 表格脚注环境
        \item[$^*$] Reported in \cite{swiftsage}
        \item[$\dagger$] Reported in \cite{subgoal_distillation}
        \item[$\ddagger$] Reported in \cite{reasonplanner}
    \end{tablenotes}
    \caption{Performance comparison between baseline frameworks and KnowMap on the ScienceWorld benchmark. All frameworks utilize \texttt{gpt-4-turbo} as their LLM backbone. Frameworks without dedicated retrieval mechanisms were provided with two randomly sampled trajectories from analogous tasks as in-context demonstrations. The \textit{"Using LLM"} column indicates whether the framework employs LLM during decision-making, while the \textit{"Tuning"} column shows if task-specific fine-tuning was applied. KnowMap receives a semi-check mark in the \textit{Tuning} column because only its knowledge embedder was fine-tuned, not the core agent model.}
    \label{tab:main_result}
    \vspace{-0.2in}
\end{threeparttable}
\end{table}

\textbf{Comparison with baselines}. 
Table.~\ref{tab:main_result} compares the scores of various agent framework. Among them, the two rows which have a check in the \textit{"Tuning"} column, \textit{SwiftSage} and \textit{Sub-goal Distillation}, tuned with a 3 billion parameter FLAN-T5-Large~\cite{flan_t5} model to manage a portion of their decision-making processes. The full version of KnowMap, in contrast, was fine-tuned with a much smaller 0.56 billion parameter embedding model, hence is given a half-check in this column. KnowMap w/o embedder tuning relies solely on KnowMap's knowledge base structure, utilizing a general embedding model for knowledge embedding and retrieval without any specific tuning.

It can be seen that KnowMap w/o embedder tuning demonstrates the best performance among all frameworks that do not undergo tuning, surpassing ReasonPlanner—another knowledge-based approach—by 2.25\%. When KnowMap's knowledge embedder is fine-tuned, its performance significantly improves, achieving a 17.20\% gain over ReasonPlanner. Although a gap of 11.06\% still exists compared to SwiftSage, KnowMap offers advantages in resource-constrained environments due to the significantly lower engineering complexity associated with its training.

% 可能考虑只画成图不画这个表了？
\begin{table}[tb]
\centering
\resizebox{0.4\textwidth}{!}{
  \begin{tabular}{c | c | c }
    \toprule 
     Backbone & Version & Score (\%) \\ \midrule
    \multirow{3}{*}{gpt-4-turbo} & \cellcolor{gray!20}few-shot & \cellcolor{gray!20}64.78 \\
    & knowledge base & 66.90 \\
    & \cellcolor{gray!20}fine-tuned embedder & \cellcolor{gray!20}76.25 \\ \hline
    \multirow{3}{*}{gpt-4o-mini} & few-shot & 49.18 \\
    & \cellcolor{gray!20}knowledge base & \cellcolor{gray!20}50.78 \\
    & fine-tuned embedder & 56.34 \\ \hline
    \multirow{3}{*}{DeepSeek-V3} & \cellcolor{gray!20}few-shot & \cellcolor{gray!20}80.50 \\
    & knowledge base & 78.84 \\
    & \cellcolor{gray!20}fine-tuned embedder & \cellcolor{gray!20}85.55 \\
    \bottomrule 
   \end{tabular}
}
    \caption{Scores with Different LLM Backbones. \textit{"few-shot"} indicates using randomly selected examples of the same task without leveraging the KnowMap knowledge base. \textit{"knowledge base"} denotes the setup utilizing the KnowMap knowledge base but without fine-tuning the knowledge embedder. \textit{"fine-tuned embedder"} refers to the complete KnowMap framework with the knowledge embedder fine-tuned.}
    \label{tab:backbone}
    \vspace{-0.4in}
\end{table}

\textbf{KnowMap's task adaptation effectiveness across LLM backbones}. KnowMap's Task Adaptation Effectiveness Across LLM Backbones
We evaluated KnowMap's effectiveness in task adaptation across various LLM backbones, as detailed in Table.~\ref{tab:backbone}. KnowMap significantly boosted performance on \texttt{gpt-4-turbo}, \texttt{gpt-4o-mini} and \texttt{DeepSeek-V3} by 17.71\%, 14.56\%, and 6.27\%, respectively. Further improvements of 13.98\%, 10.95\%, and 8.51\% were observed with embedder fine-tuning for these models, demonstrating KnowMap's broad applicability.

It's notable that for models already exhibiting strong baseline performance, such as \texttt{DeepSeek-V3}, integrating a knowledge base without fine-tuning the embedder did not yield additional improvements, even leading to slight performance degradation. This suggests that these models possess a robust inherent capability to extract and leverage knowledge directly from their original action trajectory. If the performance of the retrieval module cannot keep up with the decision model, the knowledge it provides may not be valuable enough. However, when equipped with a fine-tuned strong retrieval model, useful knowledge can still significantly improve the performance of the agent.

\begin{table}[tb]
\centering
\resizebox{0.49\textwidth}{!}{
  \begin{tabular}{c | c | c | c | c}
    \toprule 
     Method & \makecell{Pre-trained \\embedding model} & Joint knowledge &  \makecell{Fine-tuned} & Score (\%) \\ \hline
    \multirow{5}{*}{KnowMap} & no knowledge base & \cellcolor{gray!20}- & \cellcolor{gray!20}- & \cellcolor{gray!20}64.78 \\
    \cline{2-2} & \multirow{4}{*}{bge-base-env-v1.5} & \XSolidBrush & \XSolidBrush & 65.78 \\
     &  & \cellcolor{gray!20}\Checkmark & \cellcolor{gray!20}\XSolidBrush & \cellcolor{gray!20}66.90 \\
    &  & \XSolidBrush & \Checkmark & 69.11\\
     &  & \cellcolor{gray!20}\Checkmark & \cellcolor{gray!20}\Checkmark & \cellcolor{gray!20}76.25\\
    \bottomrule 
   \end{tabular}
}
    \caption{The results of ablation study. The \textit{"joint knowledge"} column indicates whether two types of knowledge were jointly utilized in retrieval, while the \textit{"fine-tuned"} column shows whether the embedding model was fine-tuned.}
    \label{tab:ablation}
    \vspace{-0.2in}
\end{table}

\textbf{The role of synergistic environmental and experiential knowledge}. To further investigate the impact of jointly leveraging environmental and experiential knowledge on KnowMap's overall performance, we conducted detailed experiments. This joint utilization of knowledge was implemented through three distinct integration approaches:

\begin{enumerate}
    \item \textbf{Integration into query}: Environmental knowledge retrieval results were directly incorporated into the input query for experiential knowledge retrieval.
    \item \textbf{Augmentation of document}: Relevant environmental knowledge was added as supplementary content within the experiential knowledge document during its derivation.
    \item \textbf{Fine-tuning protocol}: A embedding model is fine-tuned to encode both environmental and experiential knowledge. And the environmental knowledge was integrated into experiential knowledge queries during training. This ensured the model learned to intrinsically associate environmental context with experiential data.
\end{enumerate}

As illustrated in Table.~\ref{tab:ablation}, in scenarios without fine-tuning, this joint approach yielded only a marginal performance improvement of 1.07\% compared to the non-joint approach. This finding suggests that a mere combination of knowledge types solely at retrieval time—without optimizing the underlying embedding space—provides limited benefits. In stark contrast, when coupled with fine-tuning, the joint approach achieved a substantial improvement of 10.33\%. This pronounced divergence underscores that the synergistic integration of environmental and experiential knowledge primarily delivers significant performance gains when complemented by embedder fine-tuning.

Our findings suggest that while environmental knowledge provides valuable contextual signals for enhancing experiential knowledge retrieval, off-the-shelf, pre-trained embedders cannot effectively leverage these signals.  Fine-tuning is essential because it allows the model to learn robust representations that capture cross-knowledge relationships, enabling it to intrinsically associate environmental context with experiential knowledge, thereby unlocking the full synergistic potential of combining these knowledge types.

\section{CONCLUSION}

In conclusion, we proposed KnowMap, a novel approach to enhancing the adaptability of Large Language Models (LLMs) to specialized tasks. By dynamically constructing a knowledge base from environmental and experiential data and equipping LLMs with task-specific knowledge through a fine-tuned embedding model, KnowMap addresses the limitations of static pre-trained knowledge and mitigates the drawbacks of traditional fine-tuning methods. Our work demonstrates that integrating dynamic knowledge into LLMs can significantly improve their performance and versatility.

In the future, we plan to extend KnowMap's application to a wider variety of tasks beyond text-world environments. We will also explore the potential of KnowMap's knowledge construction approach in knowledge distillation and other knowledge transfer scenarios.

\bibliographystyle{ACM-Reference-Format}
\bibliography{reference}

%%% -*-BibTeX-*-
%%% Do NOT edit. File created by BibTeX with style
%%% ACM-Reference-Format-Journals [18-Jan-2012].

\begin{thebibliography}{20}

%%% ====================================================================
%%% NOTE TO THE USER: you can override these defaults by providing
%%% customized versions of any of these macros before the \bibliography
%%% command.  Each of them MUST provide its own final punctuation,
%%% except for \shownote{} and \showURL{}.  The latter two
%%% do not use final punctuation, in order to avoid confusing it with
%%% the Web address.
%%%
%%% To suppress output of a particular field, define its macro to expand
%%% to an empty string, or better, \unskip, like this:
%%%
%%% \newcommand{\showURL}[1]{\unskip}   % LaTeX syntax
%%%
%%% \def \showURL #1{\unskip}           % plain TeX syntax
%%%
%%% ====================================================================

\ifx \showCODEN    \undefined \def \showCODEN     #1{\unskip}     \fi
\ifx \showISBNx    \undefined \def \showISBNx     #1{\unskip}     \fi
\ifx \showISBNxiii \undefined \def \showISBNxiii  #1{\unskip}     \fi
\ifx \showISSN     \undefined \def \showISSN      #1{\unskip}     \fi
\ifx \showLCCN     \undefined \def \showLCCN      #1{\unskip}     \fi
\ifx \shownote     \undefined \def \shownote      #1{#1}          \fi
\ifx \showarticletitle \undefined \def \showarticletitle #1{#1}   \fi
\ifx \showURL      \undefined \def \showURL       {\relax}        \fi
% The following commands are used for tagged output and should be
% invisible to TeX
\providecommand\bibfield[2]{#2}
\providecommand\bibinfo[2]{#2}
\providecommand\natexlab[1]{#1}
\providecommand\showeprint[2][]{arXiv:#2}

\bibitem[Ahn et~al\mbox{.}(2022)]%
        {saycan}
\bibfield{author}{\bibinfo{person}{Michael Ahn}, \bibinfo{person}{Anthony Brohan}, \bibinfo{person}{Noah Brown}, \bibinfo{person}{Yevgen Chebotar}, \bibinfo{person}{Omar Cortes}, \bibinfo{person}{Byron David}, \bibinfo{person}{Chelsea Finn}, \bibinfo{person}{Chuyuan Fu}, \bibinfo{person}{Keerthana Gopalakrishnan}, \bibinfo{person}{Karol Hausman}, {et~al\mbox{.}}} \bibinfo{year}{2022}\natexlab{}.
\newblock \showarticletitle{Do as i can, not as i say: Grounding language in robotic affordances}.
\newblock \bibinfo{journal}{\emph{arXiv preprint arXiv:2204.01691}} (\bibinfo{year}{2022}).
\newblock


\bibitem[Ammanabrolu and Riedl(2021)]%
        {ammanabrolu2021learning}
\bibfield{author}{\bibinfo{person}{Prithviraj Ammanabrolu} {and} \bibinfo{person}{Mark Riedl}.} \bibinfo{year}{2021}\natexlab{}.
\newblock \showarticletitle{Learning knowledge graph-based world models of textual environments}.
\newblock \bibinfo{journal}{\emph{Advances in Neural Information Processing Systems}}  \bibinfo{volume}{34} (\bibinfo{year}{2021}), \bibinfo{pages}{3720--3731}.
\newblock


\bibitem[Anokhin et~al\mbox{.}(2024)]%
        {anokhin2024arigraph}
\bibfield{author}{\bibinfo{person}{Petr Anokhin}, \bibinfo{person}{Nikita Semenov}, \bibinfo{person}{Artyom Sorokin}, \bibinfo{person}{Dmitry Evseev}, \bibinfo{person}{Andrey Kravchenko}, \bibinfo{person}{Mikhail Burtsev}, {and} \bibinfo{person}{Evgeny Burnaev}.} \bibinfo{year}{2024}\natexlab{}.
\newblock \showarticletitle{Arigraph: Learning knowledge graph world models with episodic memory for llm agents}.
\newblock \bibinfo{journal}{\emph{arXiv preprint arXiv:2407.04363}} (\bibinfo{year}{2024}).
\newblock


\bibitem[Atzeni et~al\mbox{.}(2021)]%
        {atzeni2021case}
\bibfield{author}{\bibinfo{person}{Mattia Atzeni}, \bibinfo{person}{Shehzaad Dhuliawala}, \bibinfo{person}{Keerthiram Murugesan}, {and} \bibinfo{person}{Mrinmaya Sachan}.} \bibinfo{year}{2021}\natexlab{}.
\newblock \showarticletitle{Case-based reasoning for better generalization in textual reinforcement learning}.
\newblock \bibinfo{journal}{\emph{arXiv preprint arXiv:2110.08470}} (\bibinfo{year}{2021}).
\newblock


\bibitem[Brown et~al\mbox{.}(2020)]%
        {prompt_engnieering}
\bibfield{author}{\bibinfo{person}{Tom Brown}, \bibinfo{person}{Benjamin Mann}, \bibinfo{person}{Nick Ryder}, \bibinfo{person}{Melanie Subbiah}, \bibinfo{person}{Jared~D Kaplan}, \bibinfo{person}{Prafulla Dhariwal}, \bibinfo{person}{Arvind Neelakantan}, \bibinfo{person}{Pranav Shyam}, \bibinfo{person}{Girish Sastry}, \bibinfo{person}{Amanda Askell}, {et~al\mbox{.}}} \bibinfo{year}{2020}\natexlab{}.
\newblock \showarticletitle{Language models are few-shot learners}.
\newblock \bibinfo{journal}{\emph{Advances in neural information processing systems}}  \bibinfo{volume}{33} (\bibinfo{year}{2020}), \bibinfo{pages}{1877--1901}.
\newblock


\bibitem[Chen et~al\mbox{.}(2024)]%
        {bgem3}
\bibfield{author}{\bibinfo{person}{Jianlv Chen}, \bibinfo{person}{Shitao Xiao}, \bibinfo{person}{Peitian Zhang}, \bibinfo{person}{Kun Luo}, \bibinfo{person}{Defu Lian}, {and} \bibinfo{person}{Zheng Liu}.} \bibinfo{year}{2024}\natexlab{}.
\newblock \showarticletitle{Bge m3-embedding: Multi-lingual, multi-functionality, multi-granularity text embeddings through self-knowledge distillation}.
\newblock \bibinfo{journal}{\emph{arXiv preprint arXiv:2402.03216}} (\bibinfo{year}{2024}).
\newblock


\bibitem[Chung et~al\mbox{.}(2024)]%
        {flan_t5}
\bibfield{author}{\bibinfo{person}{Hyung~Won Chung}, \bibinfo{person}{Le Hou}, \bibinfo{person}{Shayne Longpre}, \bibinfo{person}{Barret Zoph}, \bibinfo{person}{Yi Tay}, \bibinfo{person}{William Fedus}, \bibinfo{person}{Yunxuan Li}, \bibinfo{person}{Xuezhi Wang}, \bibinfo{person}{Mostafa Dehghani}, \bibinfo{person}{Siddhartha Brahma}, {et~al\mbox{.}}} \bibinfo{year}{2024}\natexlab{}.
\newblock \showarticletitle{Scaling instruction-finetuned language models}.
\newblock \bibinfo{journal}{\emph{Journal of Machine Learning Research}} \bibinfo{volume}{25}, \bibinfo{number}{70} (\bibinfo{year}{2024}), \bibinfo{pages}{1--53}.
\newblock


\bibitem[Dinh et~al\mbox{.}(2024)]%
        {reasonplanner}
\bibfield{author}{\bibinfo{person}{Minh~Pham Dinh}, \bibinfo{person}{Munira Syed}, \bibinfo{person}{Michael~G Yankoski}, {and} \bibinfo{person}{Trenton~W Ford}.} \bibinfo{year}{2024}\natexlab{}.
\newblock \showarticletitle{ReasonPlanner: Enhancing Autonomous Planning in Dynamic Environments with Temporal Knowledge Graphs and LLMs}.
\newblock \bibinfo{journal}{\emph{arXiv preprint arXiv:2410.09252}} (\bibinfo{year}{2024}).
\newblock


\bibitem[Hashemzadeh et~al\mbox{.}(2024)]%
        {subgoal_distillation}
\bibfield{author}{\bibinfo{person}{Maryam Hashemzadeh}, \bibinfo{person}{Elias Stengel-Eskin}, \bibinfo{person}{Sarath Chandar}, {and} \bibinfo{person}{Marc-Alexandre Cote}.} \bibinfo{year}{2024}\natexlab{}.
\newblock \showarticletitle{Sub-goal Distillation: A Method to Improve Small Language Agents}.
\newblock \bibinfo{journal}{\emph{arXiv preprint arXiv:2405.02749}} (\bibinfo{year}{2024}).
\newblock


\bibitem[Huang et~al\mbox{.}(2024a)]%
        {wese}
\bibfield{author}{\bibinfo{person}{Xu Huang}, \bibinfo{person}{Weiwen Liu}, \bibinfo{person}{Xiaolong Chen}, \bibinfo{person}{Xingmei Wang}, \bibinfo{person}{Defu Lian}, \bibinfo{person}{Yasheng Wang}, \bibinfo{person}{Ruiming Tang}, {and} \bibinfo{person}{Enhong Chen}.} \bibinfo{year}{2024}\natexlab{a}.
\newblock \showarticletitle{Wese: Weak exploration to strong exploitation for llm agents}.
\newblock \bibinfo{journal}{\emph{arXiv preprint arXiv:2404.07456}} (\bibinfo{year}{2024}).
\newblock


\bibitem[Huang et~al\mbox{.}(2024b)]%
        {huang2024wese}
\bibfield{author}{\bibinfo{person}{Xu Huang}, \bibinfo{person}{Weiwen Liu}, \bibinfo{person}{Xiaolong Chen}, \bibinfo{person}{Xingmei Wang}, \bibinfo{person}{Defu Lian}, \bibinfo{person}{Yasheng Wang}, \bibinfo{person}{Ruiming Tang}, {and} \bibinfo{person}{Enhong Chen}.} \bibinfo{year}{2024}\natexlab{b}.
\newblock \showarticletitle{Wese: Weak exploration to strong exploitation for llm agents}.
\newblock \bibinfo{journal}{\emph{arXiv preprint arXiv:2404.07456}} (\bibinfo{year}{2024}).
\newblock


\bibitem[Jiang et~al\mbox{.}(2024)]%
        {jiang2024kg}
\bibfield{author}{\bibinfo{person}{Jinhao Jiang}, \bibinfo{person}{Kun Zhou}, \bibinfo{person}{Wayne~Xin Zhao}, \bibinfo{person}{Yang Song}, \bibinfo{person}{Chen Zhu}, \bibinfo{person}{Hengshu Zhu}, {and} \bibinfo{person}{Ji-Rong Wen}.} \bibinfo{year}{2024}\natexlab{}.
\newblock \showarticletitle{Kg-agent: An efficient autonomous agent framework for complex reasoning over knowledge graph}.
\newblock \bibinfo{journal}{\emph{arXiv preprint arXiv:2402.11163}} (\bibinfo{year}{2024}).
\newblock


\bibitem[Lin et~al\mbox{.}(2023)]%
        {swiftsage}
\bibfield{author}{\bibinfo{person}{Bill~Yuchen Lin}, \bibinfo{person}{Yicheng Fu}, \bibinfo{person}{Karina Yang}, \bibinfo{person}{Faeze Brahman}, \bibinfo{person}{Shiyu Huang}, \bibinfo{person}{Chandra Bhagavatula}, \bibinfo{person}{Prithviraj Ammanabrolu}, \bibinfo{person}{Yejin Choi}, {and} \bibinfo{person}{Xiang Ren}.} \bibinfo{year}{2023}\natexlab{}.
\newblock \showarticletitle{Swiftsage: A generative agent with fast and slow thinking for complex interactive tasks}.
\newblock \bibinfo{journal}{\emph{Advances in Neural Information Processing Systems}}  \bibinfo{volume}{36} (\bibinfo{year}{2023}), \bibinfo{pages}{23813--23825}.
\newblock


\bibitem[Shinn et~al\mbox{.}(2023)]%
        {reflexion}
\bibfield{author}{\bibinfo{person}{Noah Shinn}, \bibinfo{person}{Federico Cassano}, \bibinfo{person}{Ashwin Gopinath}, \bibinfo{person}{Karthik Narasimhan}, {and} \bibinfo{person}{Shunyu Yao}.} \bibinfo{year}{2023}\natexlab{}.
\newblock \showarticletitle{Reflexion: Language agents with verbal reinforcement learning}.
\newblock \bibinfo{journal}{\emph{Advances in Neural Information Processing Systems}}  \bibinfo{volume}{36} (\bibinfo{year}{2023}), \bibinfo{pages}{8634--8652}.
\newblock


\bibitem[Song et~al\mbox{.}(2023)]%
        {song2023llm}
\bibfield{author}{\bibinfo{person}{Chan~Hee Song}, \bibinfo{person}{Jiaman Wu}, \bibinfo{person}{Clayton Washington}, \bibinfo{person}{Brian~M Sadler}, \bibinfo{person}{Wei-Lun Chao}, {and} \bibinfo{person}{Yu Su}.} \bibinfo{year}{2023}\natexlab{}.
\newblock \showarticletitle{Llm-planner: Few-shot grounded planning for embodied agents with large language models}. In \bibinfo{booktitle}{\emph{Proceedings of the IEEE/CVF international conference on computer vision}}. \bibinfo{pages}{2998--3009}.
\newblock


\bibitem[Tziafas and Kasaei(2024)]%
        {LRLL}
\bibfield{author}{\bibinfo{person}{Georgios Tziafas} {and} \bibinfo{person}{Hamidreza Kasaei}.} \bibinfo{year}{2024}\natexlab{}.
\newblock \showarticletitle{Lifelong robot library learning: Bootstrapping composable and generalizable skills for embodied control with language models}. In \bibinfo{booktitle}{\emph{2024 IEEE International Conference on Robotics and Automation (ICRA)}}. IEEE, \bibinfo{pages}{515--522}.
\newblock


\bibitem[Wang et~al\mbox{.}(2022)]%
        {scienceworld}
\bibfield{author}{\bibinfo{person}{Ruoyao Wang}, \bibinfo{person}{Peter Jansen}, \bibinfo{person}{Marc-Alexandre C{\^o}t{\'e}}, {and} \bibinfo{person}{Prithviraj Ammanabrolu}.} \bibinfo{year}{2022}\natexlab{}.
\newblock \showarticletitle{Scienceworld: Is your agent smarter than a 5th grader?}
\newblock \bibinfo{journal}{\emph{arXiv preprint arXiv:2203.07540}} (\bibinfo{year}{2022}).
\newblock


\bibitem[Wei et~al\mbox{.}(2022)]%
        {wei2022chain}
\bibfield{author}{\bibinfo{person}{Jason Wei}, \bibinfo{person}{Xuezhi Wang}, \bibinfo{person}{Dale Schuurmans}, \bibinfo{person}{Maarten Bosma}, \bibinfo{person}{Fei Xia}, \bibinfo{person}{Ed Chi}, \bibinfo{person}{Quoc~V Le}, \bibinfo{person}{Denny Zhou}, {et~al\mbox{.}}} \bibinfo{year}{2022}\natexlab{}.
\newblock \showarticletitle{Chain-of-thought prompting elicits reasoning in large language models}.
\newblock \bibinfo{journal}{\emph{Advances in neural information processing systems}}  \bibinfo{volume}{35} (\bibinfo{year}{2022}), \bibinfo{pages}{24824--24837}.
\newblock


\bibitem[Yao et~al\mbox{.}(2023)]%
        {react}
\bibfield{author}{\bibinfo{person}{Shunyu Yao}, \bibinfo{person}{Jeffrey Zhao}, \bibinfo{person}{Dian Yu}, \bibinfo{person}{Nan Du}, \bibinfo{person}{Izhak Shafran}, \bibinfo{person}{Karthik Narasimhan}, {and} \bibinfo{person}{Yuan Cao}.} \bibinfo{year}{2023}\natexlab{}.
\newblock \showarticletitle{React: Synergizing reasoning and acting in language models}. In \bibinfo{booktitle}{\emph{International Conference on Learning Representations (ICLR)}}.
\newblock


\bibitem[Zhao et~al\mbox{.}(2024)]%
        {zhao2024expel}
\bibfield{author}{\bibinfo{person}{Andrew Zhao}, \bibinfo{person}{Daniel Huang}, \bibinfo{person}{Quentin Xu}, \bibinfo{person}{Matthieu Lin}, \bibinfo{person}{Yong-Jin Liu}, {and} \bibinfo{person}{Gao Huang}.} \bibinfo{year}{2024}\natexlab{}.
\newblock \showarticletitle{Expel: Llm agents are experiential learners}. In \bibinfo{booktitle}{\emph{Proceedings of the AAAI Conference on Artificial Intelligence}}, Vol.~\bibinfo{volume}{38}. \bibinfo{pages}{19632--19642}.
\newblock


\end{thebibliography}
\end{document}